\documentclass[conference,10pt]{IEEEtran}
\usepackage{epsfig,rotating,setspace,latexsym,amsmath,epsf,amssymb,amsfonts,bm,theorem,subfigure,epstopdf}
\usepackage{cite,authblk}
\usepackage{bbm}
\usepackage{color}
\usepackage{mathtools}
\usepackage{algorithm}
\usepackage[noend]{algpseudocode}
\usepackage{comment}

\algrenewcommand\algorithmicforall{\textbf{foreach}}
\algrenewcommand\algorithmicindent{.8em}

\DeclareMathOperator*{\argmin}{arg\,min}
\DeclareMathOperator*{\otil}{\tilde{\omega}}

\IEEEoverridecommandlockouts
\allowdisplaybreaks

\begin{document}

\title{FedGradNorm: Personalized Federated Gradient-Normalized Multi-Task Learning}
\author{Matin Mortaheb \qquad Cemil Vahapoglu  \qquad Sennur Ulukus\\
\normalsize Department of Electrical and Computer Engineering\\
\normalsize University of Maryland, College Park, MD 20742\\
\normalsize  \emph{mortaheb@umd.edu} \qquad \emph{cemilnv@umd.edu}  \qquad \emph{ulukus@umd.edu}}
	
\maketitle

\begin{abstract}
Multi-task learning (MTL) is a novel framework to learn several tasks simultaneously with a single shared network where each task has its distinct personalized header network for fine-tuning. MTL can be implemented in federated learning settings as well, in which tasks are distributed across clients. In federated settings, the statistical heterogeneity due to different task complexities and data heterogeneity due to non-iid nature of local datasets can both degrade the learning performance of the system. In addition, tasks can negatively affect each other's learning performance due to negative transference effects. To cope with these challenges, we propose \emph{FedGradNorm} which uses a dynamic-weighting method to normalize gradient norms in order to balance learning speeds among different tasks. \emph{FedGradNorm} improves the overall learning performance in a personalized federated learning setting. We provide convergence analysis for \emph{FedGradNorm} by showing that it has an exponential convergence rate. We also conduct experiments on multi-task facial landmark (MTFL) and wireless communication system dataset (RadComDynamic). The experimental results show that our framework can achieve faster training performance compared to equal-weighting strategy. In addition to improving training speed, \emph{FedGradNorm} also compensates for the imbalanced datasets among clients.
\end{abstract}
 
\section{Introduction}

Multi-task learning (MTL) is a learning paradigm that aims to learn multiple related tasks simultaneously by learning a shared representation for all tasks \cite{MTLintro, Zhang2017ASO}. MTL is motivated by the idea that different data used for related tasks can have a common representation \cite{Bengio2013RepresentationLA}. In MTL, all tasks share the pre-layers of a network, and task-specific layers are stacked on top of the shared base to output task-specific predictions. The loss of the overall model is expressed as a weighted sum of individual task losses multiplied by their corresponding loss weights. Using the synergy among multiple tasks, MTL offers data efficiency, robust regularization through a shared representation, better overall system performance, and fast learning exploiting the auxiliary information \cite{Crawshaw2020MultiTaskLW}. MTL is particularly suitable for distributed learning settings, as it may not be feasible to expect a single centralized unit to have data and labels relevant to multiple fundamentally different tasks. Thus, in this paper, we consider MTL in a federated learning setting, enhanced with personalization.

\begin{figure}[t]
 \centerline{\includegraphics[width=0.9\linewidth]{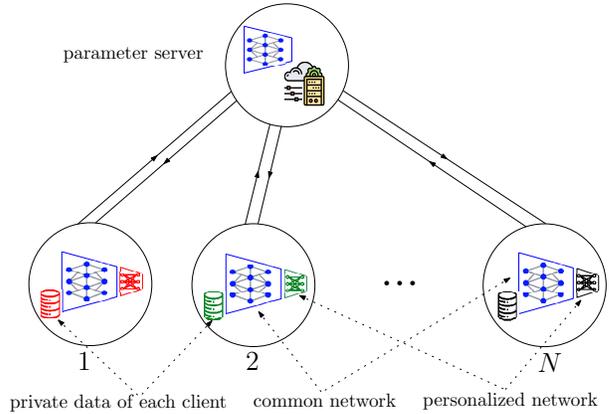}}
  \caption{Personalized federated learning framework with a common network (shown in blue) and small personalized headers (shown in red, green, black).}
  \label{FLgrad}
  \vspace*{-0.4cm}
\end{figure}

Federated learning (FL) is a distributed learning framework where many clients train a shared model under the orchestration of a centralized server while keeping the training data decentralized and private. In personalized federated learning (PFL), see Fig.~\ref{FLgrad}, clients have different tasks: while the parameter server and the clients train a common base model, each client further trains a small header network for its own specific task, referred to as personalization. PFL enables essentially different learning models in individual clients that better fit user-specific data while also capturing the common knowledge distilled from data of other devices \cite{collins2021exploiting, arivazhagan2019federated, deng2020adaptive}. The most relevant PFL works for our paper are federated representation learning (\emph{FedRep}) \cite{collins2021exploiting} and federated learning with personalization layers (\emph{FedPer}) \cite{arivazhagan2019federated}. In both \emph{FedPer} and \emph{FedRep}, the clients share data representation across the shared global network, and use unique task-specific local heads. Further, in order to maintain fair training among multiple tasks, cope with statistical heterogeneity due to task complexity and data distribution, and suppress negative transference effects between tasks, adaptive weighting strategies can be used to adjust the weights of the task losses over time. \emph{GradNorm} \cite{chen2018gradnorm} is a dynamic weighting algorithm that scales the task loss functions with respect to the learning speeds in order to normalize the gradient norms. 

\emph{FedPer} \cite{arivazhagan2019federated} and 
\emph{FedRep} \cite{collins2021exploiting}
consider only the equal-weights case to aggregate the clients' loss, and \emph{GradNorm} \cite{chen2018gradnorm} considers dynamic weights in MTL, but does not consider a distributed FL setting. We combine some aspects of \cite{arivazhagan2019federated, collins2021exploiting, chen2018gradnorm} to create our novel framework called \emph{FedGradNorm}, where we incorporate dynamic-weighting inside a PFL setting. We provide theoretical convergence proof for our framework, while \emph{FedPer} \cite{arivazhagan2019federated} and \emph{GradNorm} \cite{chen2018gradnorm} do not provide any convergence proofs, and \emph{FedRep} \cite{collins2021exploiting} provides a convergence proof only for the special case of linear personalized header. The experimental and theoretical results show the superiority of our framework to the equal-weighting PFL methods \cite{arivazhagan2019federated, collins2021exploiting}.

The main contributions of our paper are as follows:
i) We propose \emph{FedGradNorm} algorithm, which uses \emph{GradNorm} dynamic-weighting strategy in a PFL setup to achieve a better and fair learning performance when clients have different tasks. ii) We provide convergence analysis for our \emph{FedGradNorm} algorithm utilizing \emph{bilevel optimization} \cite{sinha2017review,ji2021bilevel}. To the best of our knowledge, this is the first work that provides convergence analysis for \emph{GradNorm} adaptive-weighting strategy in addition to applying it in an FL setting. iii) We conduct several experiments on our framework using multi-task facial landmark (MTFL) dataset \cite{zhang2014facial}, and RadComDynamic wireless communications dataset \cite{JagannathMTL}. To compare the performance of the learning speed of \emph{FedGradNorm} with equal-weighting method, we investigate the change in the task loss during the training phase. Experimental results exhibit better performance in \emph{FedGradNorm} than \emph{FedRep}. 

\section{System Model and Problem Formulation}

\subsection{Federated Learning (FL) Setup}
The generic form of FL problem with $N$ clients is
\begin{align}
        \min_{\omega} \{ F(\omega) \triangleq \frac{1}{N} \sum_{i=1}^N p^{(i)}F^{(i)}(\omega)\}
\end{align}
where $p^{(i)}$ is the loss weight for client $i$ such that $\sum_{i=1}^N p^{(i)}= N $, and $F^{(i)}$ is the local loss function for client $i$.

\subsection{Personalized Federated Multi-Task Learning (PF-MTL)}
We consider a PFL setting with $N$ clients, in which client $i$ has its own local dataset $D_i = \{(\mathbf{x}^{(i)}_j,y^{(i)}_j) \}_{j=1}^{n_i}$ where $n_i$ is the size of the local dataset. $T_i$ denotes the task of client $i$, $\forall i \in [N]$. The system model consists of a global representation network $q_\omega :\mathbb{R}^{d} \rightarrow \mathbb{R}^{d'}$ which is a function parameterized by $\omega \in \mathcal{W}$ and maps data points to a lower space of size $d'$. All clients share the same global representation network which is synchronized across clients with global aggregation. The client-specific heads $q_{h^{(i)}} : \mathbb{R}^{d'} \rightarrow \mathcal{Y}$ are functions parameterized by $h^{(i)} \in \mathcal{H}$ for all clients $i \in [N]$ and map from the low dimensional representation space to the label space $\mathcal{Y}$. The system model is shown in Fig.~\ref{FLgrad}. The local model for client $i$ is the composition of the $i$th client's global representation model $q_\omega$ and  personalized model $q_{h^{(i)}}$,  shown as $q_i (\cdot) = (q_{h^{(i)}} \circ q_\omega)(\cdot)$. In addition, the local loss for the $i$th client is $F^{(i)}(h^{(i)},\omega) = F^{(i)}(q_i (\cdot)) = F^{(i)}((q_{h^{(i)}} \circ q_\omega)(\cdot))$.

The clients and the centralized server aim at learning global representation parameters $\omega$ together, while each client $i$ learns its unique client-specific parameters $h^{(i)}$ locally by performing alternating minimization. Specifically, client $i$ performs $\tau_h$ local gradient based updates to optimize $h^{(i)}$ while global network parameters at client $i$, i.e., $ \omega^{(i)}$ is frozen. Thereafter, client $i$ performs $\tau_\omega$ local updates for optimizing the global shared network parameters at client $i$ while the parameters corresponding to the client-specific head are frozen. Then, the global shared network parameters $\{\omega^{(i)} \}_{i=1}^N$ are aggregated at the centralized server to have a common $\omega$. Thus, the optimization problem is
\begin{align}\label{min-min}
    \min_{\omega \in \mathcal{W}} \frac{1}{N} \sum_{i=1}^N p^{(i)} \min_{h^{(i)} \in \mathcal{H}} F^{(i)}(h^{(i)},\omega)
\end{align}
\emph{FedRep} \cite{collins2021exploiting} investigates this problem for $p^{(i)}=1$ for $\forall i \in [N]$. 

\subsection{PF-MTL as Bilevel Optimization Problem }
Since $F^{(i)}(h^{(i)},\omega)$ depends on only $h^{(i)}$ and $\omega$ for all $i \in [N]$, we can rewrite (\ref{min-min}) as
\begin{align} \label{final_opt}
    \min_{\omega \in \mathcal{W} , \{h^{(i)} \in \mathcal{H}\}_{i=1}^N} \frac{1}{N} \sum_{i=1}^N p^{(i)} F^{(i)}(h^{(i)},\omega)
\end{align}
The problem given in (\ref{final_opt}) depends on $p^{(i)}$ values obtained by our \emph{FedGradNorm} algorithm, which will be described later, as a result of another optimization problem. Thus, the problem can be written in the form of a bilevel optimization problem which is an optimization problem that contains another optimization problem as a constraint in the following form
\begin{align}
    & \min_{x_u \in X_u , x_l \in X_l}  \quad F(x_u,x_l) \nonumber \\
    & \textrm{s.t.}  \quad x_l = \argmin_{x_l \in X_l} \{ g(x_u,x_l)  : c_j(x_u,x_l) \leq 0, \; j=1,\ldots,J \} \nonumber \\
    & \qquad C_m(x_u,x_l)\leq 0, \quad m=1,\ldots,M 
\end{align}    
where $F(x_u,x_l)$ is the upper-level objective function and $g(x_u,x_l)$ is the lower-level objective function; $\{c_j(x_u,x_l) \leq 0, j=1,\ldots,J \}$ represent the constraints for the lower-level optimization problem; and $\{C_m(x_u,x_l)\leq 0, m=1,\ldots,M\}$ and the lower-level optimization problem itself represent the constraints for the upper-level optimization problem.

We utilize the iterative differentiation (ITD) algorithm \cite{ji2021bilevel} which is given in Algorithm~\ref{alg:itd}. The upper-level optimization update is performed in the outer loop, while the lower-level optimization update is performed in the inner loop. 

\begin{algorithm}[h]
    \caption{Iterative Differentiation (ITD) Algorithm.}
    \label{alg:itd}
\begin{algorithmic}
\State {\bfseries Input:} $K$,$D$, step sizes $\alpha$, $\beta$, initialization $x_u (0)$, $x_l (0)$.
\For{ $k$ = 0, 1, 2, \ldots, $K$}
\State Set $x_{l}^{0}(k)$ = $x_{l}^{D}(k-1)$ if $k > 0$ otherwise $x_l (0)$.
\For{$t$ = 1, \ldots, $D$}
\State Update $x_{l}^{t} (k) = x_{l}^{t-1} (k) -\alpha \nabla_{x_{l}} g(x_{u}(k),x_{l}^{t-1}(k))$ 
\EndFor
\State Compute $\Hat{\nabla}_{x_u} F(x_u (k), x_{l}^{D} (k) ) = \frac{\partial F(x_u (k), x_{l}^{D} (k))}{\partial x_u}$
\State Update $x_u(k+1) = x_u(k) - \beta \Hat{\nabla}_{x_u} F(x_u (k), x_{l}^{D} (k))$
\EndFor
\end{algorithmic}
\end{algorithm}

For our problem, $x_u$ and $x_l$ correspond to $\left(\{h^{(i)}\}_{i=1}^N,\omega \right)$, $\{p^{(i)}\}_{i=1}^N$, respectively. Additionally, $x_u(k)$ and $x_l(k)$ are denoted as $\left(\{h^{(i)}\}_{i=1}^N,\omega_k \right)$, $\{p^{(i)}_k\}_{i=1}^N$ to represent the outer loop iteration index in Algorithm \ref{alg:itd} for the rest of the paper. Also, $i$ in $p^{i}_k$ represents the inner loop iteration index, while $i$ in $p^{(i)}_k$ represents the client index. Then, the bilevel optimization problem in our case can be written as
\begin{align} \label{final_optimization}
    &\min_{\omega, \{h^{(i)}\}_{i=1}^N, \{p^{(i)}\}_{i=1}^N} \quad F(\{h^{(i)}\}_{i=1}^N, \omega, \{p^{(i)}\}_{i=1}^N) \nonumber \\
     &\textrm{s.t.} \quad \{p^{(i)}\}_{i=1}^N \in \argmin_{\{p^{(i)}\}_{i=1}^N \in \mathbb{R}^N} F_{grad}
\end{align}    
where our objective function is the weighted sum of the local loss functions, i.e., $F=\frac{1}{N} \sum_{i=1}^N p^{(i)} F^{(i)}(h^{(i)},\omega)$ and $F_{grad}$ is the auxiliary loss function defined in the \emph{FedGradNorm} algorithm in the next section.

\section{\emph{FedGradNorm}: Federated GradNorm Algorithm}
\emph{FedGradNorm} is a distributed dynamic weighting strategy which is implemented in an FL setup under the orchestration of a parameter server. \emph{FedGradNorm} is a generalization of \emph{GradNorm} proposed in \cite{chen2018gradnorm} in a centralized learning model.

\subsection{Definitions and Preliminaries}
In \emph{FedGradNorm}, we aim to learn the dynamic loss weights $\{p^{(i)}\}_{i=1}^N$ given in the lower-level optimization problem of (\ref{final_optimization}). The main objective of the algorithm is to dynamically adjust the gradient norms so that the different tasks across clients can be trained at similar learning speeds. In the rest of the paper, \emph{clients} and \emph{tasks} will be used interchangeably as we assume that each client has its own different task. Before describing the algorithm in detail, we first introduce the notations: 
\begin{itemize}
    \item $\tilde{\omega}$: A subset of the global shared network parameters $\tilde{\omega} \subset \omega$. \emph{FedGradNorm} is applied on $\tilde{\omega}^{(i)}_k$ $\subset$ $\omega^{(i)}_k$, which is a subset of the global shared network parameters at client $i$ at iteration $k$. $\tilde{\omega}^{(i)}_k$ is generally chosen as the last layer of the global shared network at client $i$ at iteration $k$.
    \item $G_{\tilde{\omega}^{(i)}_k}^{(i)}(k) = \| \nabla_{\otil^{(i)}_k} p^{(i)}_k F^{(i)}_k \| = p^{(i)}_k  \| \nabla_{\otil^{(i)}_k} F^{(i)}_k \|$: The $\ell_2$ norm of the gradient of the weighted task loss at client $i$ at iteration $k$ with respect to the chosen weights $\tilde{\omega}^{(i)}_k$.
    \item $\bar{G}_{\otil} (k)$ = $\mathbb{E}_{j \sim \textrm{task}} [G_{\otil^{(j)}_k}^{(j)}(k)]$: The average gradient norm across all clients (tasks) at iteration $k$.
    \item $\tilde{F}^{(i)}_k$ = $\frac{F^{(i)}_k}{F^{(i)}_0}$: Inverse training rate of task $i$ (at client $i$) at iteration $k$, where $F^{(i)}_k$ is the loss for client $i$ at iteration $k$, and $F^{(i)}_0$ is the initial loss for client $i$.
    \item $r^{(i)}_k$ =$\frac{\tilde{F}^{(i)}_k}{\mathbb{E}_{j \sim \textrm{task}}[\tilde{F}^{(j)}_k]}$: Relative inverse training rate of task $i$ at iteration $k$.
\end{itemize}
Additional notations that are useful in algorithm description:
\begin{itemize}
    \item $g_k^{(i)} = \frac{1}{\tau_\omega} \sum_{j=1}^{\tau_\omega} g_{k,j}^{(i)}$ is the average of gradient updates at client $i$ at iteration $k$, where $g_{k,j}^{(i)}$ is the $j$th local update of the global shared representation at client $i$ at iteration $k$. Note that $ \| \nabla_{\otil^{(i)}_k} F^{(i)}_k \|$ is a subset of $g_k^{(i)}$ since $\tilde{\omega} \subset \omega$.
    \item $h_{k,j}^{(i)}$ is the client-specific head parameters $h^{(i)}$ after the $j$th local update on the client-specific network of client $i$ at iteration $k$, $j = 1,\ldots, \tau_h$.
    \item $\omega_{k,j}^{(i)}$ is the global shared network parameters of client $i$ after the $j$th local update at iteration $k$, $j = 1,\ldots, \tau_\omega$. Additionally, $\omega_{k}^{(i)}$ denotes $\omega_{k,\tau_\omega}^{(i)}$ for brevity.
\end{itemize}

\subsection{FedGradNorm Description}
\emph{FedGradNorm} is used to balance the training rates of different tasks across clients by adjusting the gradient magnitudes as in the \emph{GradNorm} \cite{chen2018gradnorm}. Unlike \emph{GradNorm}, \emph{FedGradNorm} is distributed across clients and the parameter server. $\bar{G}_{\otil}$ is used to have a common scale for the gradient sizes while the gradient norms are adjusted according to the relative inverse training rates  $r^{(i)}_k$. With a higher value of $r^{(i)}_k$, a higher gradient magnitude is used for task $i$ in order to encourage the task to train more quickly. $r^{(i)}_k$ is calculated by the parameter server by using $\tilde{F}^{(i)}_k$ coming from clients. Therefore, by using the common scale of gradient magnitudes, and the relative inverse training rate, the desired gradient norm of task $i$ at iteration $k$ is determined as $\bar{G}_{\otil} (k) \times \left[ r^{(i)}_k  \right]^\gamma$, where $\gamma$ represents the strength of the restoring force which pulls tasks back to a common training rate, which can also be thought of as a metric of task asymmetry across different tasks. If tasks have different learning complexities, i.e., different learning dynamics, a larger $\gamma$ should be used for a stronger balancing.

Since we want the gradient norms to shift towards the desired gradient norm, loss weights $p^{(i)}_k$ are updated by the minimization of an auxiliary loss function $F_{\textrm{grad}} \left(k ; \{p^{(i)}_k\}_{i=1}^N \right)$ defined as the summation of $\ell_2$ distance between the actual gradient norm and the desired gradient norm across all tasks for each iteration $k$, i.e.,
\begin{align}\label{F_grad}
    & F_{\textrm{grad}} \left(k ; \{p^{(i)}_k\}_{i=1}^N \right) = \sum_{i=1}^N F^{(i)}_{\textrm{grad}} \left(k ; p^{(i)}_k\right) \nonumber\\
    & \qquad \quad = \sum_{i=1}^N  \left \| p^{(i)}_k  \| \nabla_{\otil^{(i)}_k} F^{(i)}_k \| - \bar{G}_{\otil} (k) \times [r^{(i)}_k]^\gamma \right \|
\end{align}    

The auxiliary loss function $F_{\textrm{grad}} \left(k ; \{p^{(i)}_k\}_{i=1}^N \right)$ is constructed by the parameter server at each global iteration $k$ by using $\nabla_{\otil^{(i)}_k} F^{(i)}_k$, which is a subset of the whole gradient of the global shared network sent by client $i$ at iteration $k$ for the global aggregation. In addition, clients send $\tilde{F}^{(i)}_k$ to the parameter server, so that the parameter server can construct $r^{(i)}_k$ to have the desired gradient norm. 

Next, the parameter server performs the differentiation of $F_{\textrm{grad}} \left(k ; \{p^{(i)}_k\}_{i=1}^N \right)$ with respect to each element of $\{p^{(i)}\}_{i=1}^N$ so that $\nabla_{p^{(i)}}F_{\textrm{grad}}$ is applied via gradient descent to update $p^{(i)}$. The desired gradient norm terms, $\bar{G}_{\otil} (k) \times \left[ r^{(i)}_k  \right]^\alpha$, are treated as constant to prevent loss weights $\{p^{(i)}\}_{i=1}^N$ from drifting towards zero while differentiating $F_{\textrm{grad}} \left(k ; \{p^{(i)}_k\}_{i=1}^N \right)$ with respect to each loss weight $p_k^{(i)}$. The weights are updated as,
\begin{align}
    p^{(i)} \leftarrow p^{(i)}-\alpha\nabla_{p^{(i)}}F_{\textrm{grad}}, \quad \forall i \in [N].
\end{align}

The updated $\{ p^{(i)} \}_{i=1}^N$ are normalized so that $\sum_{i=1}^N p^{(i)} = N$. Finally, the parameter server obtains the global aggregated gradient $g_k = \frac{1}{N} \sum_{i=1}^N p_k^{(i)} g^{(i)}_k$ to update the global shared network parameters $\omega$ via $\omega_{k+1} = \omega_{k} - \beta g_k$ and broadcasts the updated parameters to the clients for the next iteration. The overall \emph{FedGradNorm} algorithm is summarized in Algorithm~\ref{alg:fedgradnorm}. In \emph{FedGradNorm}, $\textrm{Update}(f,h)$ represents the generic notation for the update of the variable $h$ by using the gradient of $f$ function with respect to the variable $h$.
 
\begin{algorithm}[t]
    \caption{Training with \emph{FedGradNorm}}
    \label{alg:fedgradnorm}
\begin{algorithmic}
    \State Initialize $\omega_0$, $\{p_0^{(i)}\}_{i=1}^N$, $\{h_0^{(i)}\}_{i=1}^N$
    \For {$k$=1 {\bfseries to} $K$} 
        \State The parameter server sends the current global shared network parameters $\omega_k$ to the clients.
        \For {Each client $i \in [N]$} 
            \State Initialize global shared network parameters for local updates by $\omega_{k,0}^{(i)} \leftarrow \omega_k$
            \For{$j=1,\ldots,\tau_h$} 
                \State $h_{k,j}^{(i)}$ = $\textrm{Update} (F^{(i)}(h_{k,j-1}^{(i)}, \omega_{k,0}^{(i)}),h_{k,j-1}^{(i)})$
            \EndFor
            \State $F_k^{(i)}=0$
            \For{$j=1,\ldots,\tau_\omega$} 
                \State $\omega_{k,j}^{(i)} \leftarrow \omega_{k,j-1}^{(i)} - \beta g_{k,j}^{(i)}$
                \State $F_k^{(i)}$ += $F^{(i)}(h_{k,\tau_h}^{(i)},\omega_{k,j}^{(i)})$
            \EndFor
            \State $F_k^{(i)}$ $\leftarrow \frac{1}{\tau_\omega} F_k^{(i)} $
            \State Client $i$ sends $g_k^{(i)} = \frac{1}{\tau_\omega} \sum_{j=1}^{\tau_\omega} g_{k,j}^{(i)}$, and $\tilde{F}_k^{(i)} = \frac{F_k^{(i)}}{F_0^{(i)}}$ to the parameter server
        \EndFor
        \State After collecting $g_k^{(i)}$, and $\tilde{F}_k^{(i)}$ for active clients $i \in [N]$, the parameter server performs the following operations in the order:
            \State \textbullet~ Constructs $F_{\textrm{grad}} \left(k ; \{p^{(i)}_k\}_{i=1}^N \right)$ using $\{g_k^{(i)}\}_{i=1}^N$ and $\{\tilde{F}_k^{(i)}\}_{i=1}^N$ as given in eq. (\ref{F_grad}).
            \State \textbullet~ Updates $p_k^{(i)} \leftarrow p_{k-1}^{(i)} - \alpha \nabla_{p^{(i)}} F_{\textrm{grad}}$, $\forall i \in [N]$.
            \State \textbullet~ Aggregates the gradient for the global shared network by $g_k = \frac{1}{N} \sum_{i=1}^N  p_k^{(i)}g_k^{(i)}$.
            \State \textbullet~ Updates the global shared network parameters with the aggregated gradient by $\omega_{k+1} = \omega_{k} - \beta g_k$.
            \State \textbullet~ Broadcasts $\omega_{k+1}$ to clients for the next global iteration.
    \EndFor
\end{algorithmic}
\end{algorithm}

\subsection{Convergence of FedGradNorm}
In the convergence analysis, we assume strong convexity of the upper-level objective function $F(\cdot)$ and the lower-level objective function $F_{\textrm{grad}}(\cdot)$, the Lipschitzness of both objective functions and the Lipschitzness of the first and second order gradients of both objective functions. We prove the exponential convergence of \emph{FedGradNorm} under these assumptions. Due to space limitations, we skip the details of the proof here, and present them in the longer version. Instead, here, we present our experimental results in the next section.

\section{Experimental Results}
We compare the task losses achieved by equal-weighting in \emph{FedRep} and  dynamic-weighting in our \emph{FedGradNorm}. 

\subsection{Dataset Specifications}
We use the following two dataset for our experiments:

\noindent
\emph{Multi-task facial landmark (MTFL)} \cite{zhang2014facial} contains 10,000 training data and 3,000 test images, which are face images annotated by 1) five facial landmarks, 2) gender, 3) smiling or not, 4) wearing glasses or not, and 5) head pose.
   
\noindent 
\emph{Wireless dataset (RadComDynamic)} \cite{JagannathMTL} is a multi-class wireless signal dataset of 125,000 samples. Samples are radar and communication signals of varying SNR values from GNU radio companion. It contains 6 modulation types and 8 signal types. We perform 3 different tasks: 1) modulation classification, 2) signal type classification, and 3) anomaly detection. The modulation classes are amdsb, amssb, ask, bpsk, fmcw, pulsed continous wave (PCW). The signal type classes are AM radio, short-range, radar-altimeter, air-ground-MTI, airborne-detection, airborne-range, ground-mapping. As an anomaly behavior, we consider having an SNR lower than -4 dB since SNR can be a proxy for geo-location information, and low SNR may indicate a signal coming from an outsider. Each data point in this dataset is a normalized signal vector of size 256 obtained by vectorizing the real and complex parts of the signal ($x = x_I + jx_Q$ where $x_I,x_Q \in \mathcal{R}^{128}$).

\subsection{Hyperparameters and Model Specifications}
We choose $\gamma$ as 0.9 through our experiments. Note that $\gamma$ is the only hyperparameter of \emph{FedGradNorm}, and it should be determined with respect to the task asymmetry in the system. The learning rate $\beta$, which is used for training of global shared network and the personalized network on the client side is 0.0002, and the learning rate  $\alpha$ for $F_{\textrm{grad}}$ optimization is 0.004. We use Adam optimizer for both network training and $F_{\textrm{grad}}$ optimization. The shared network model is explained in Table~\ref{network_model}. Each client also has a simple linear layer that maps the shared network's output to the corresponding prediction value for a personalized network. Cross-entropy and mean squared error (MSE) are used as the loss functions for classification and regression tasks, respectively.

\begin{table}[h!]
\begin{center}
\begin{sc}
\begin{tabular}{|c|c|}
\hline
Network 1 & network 2 \\
\hline
Conv2d(1, 16, 5)    & FC(256, 512)\\
MaxPool2d(2, 2)  & FC(512, 1024)\\
Conv2d(16, 48, 3)   & FC(1024, 2048)\\
MaxPool2d(2, 2)    & FC(2048, 512)\\
Conv2d(48, 64, 3)  & FC(512, 256)\\
MaxPool2d(2, 2)    & \\
Conv2d(64, 64, 2)   & \\
\hline
\end{tabular}
\end{sc}
\end{center}
\caption{Shared network model.}
\label{network_model}
\vspace*{-0.5cm}
\end{table}

\subsection{Results and Analysis}

\begin{figure}[]
 	\begin{center}
 	\subfigure[]{%
 	\includegraphics[width=0.49\linewidth]{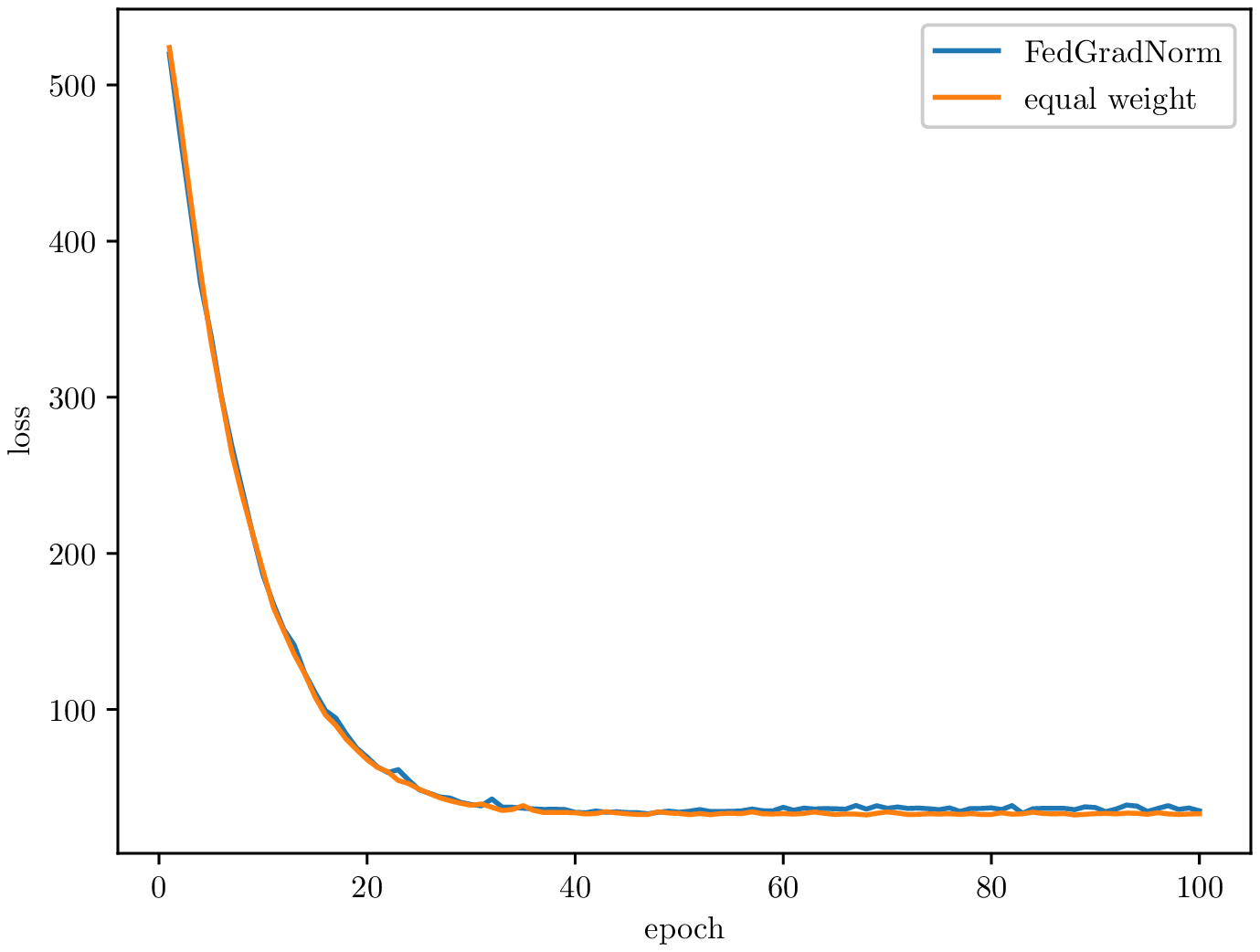}}
 	\subfigure[]{%
 	\includegraphics[width=0.49\linewidth]{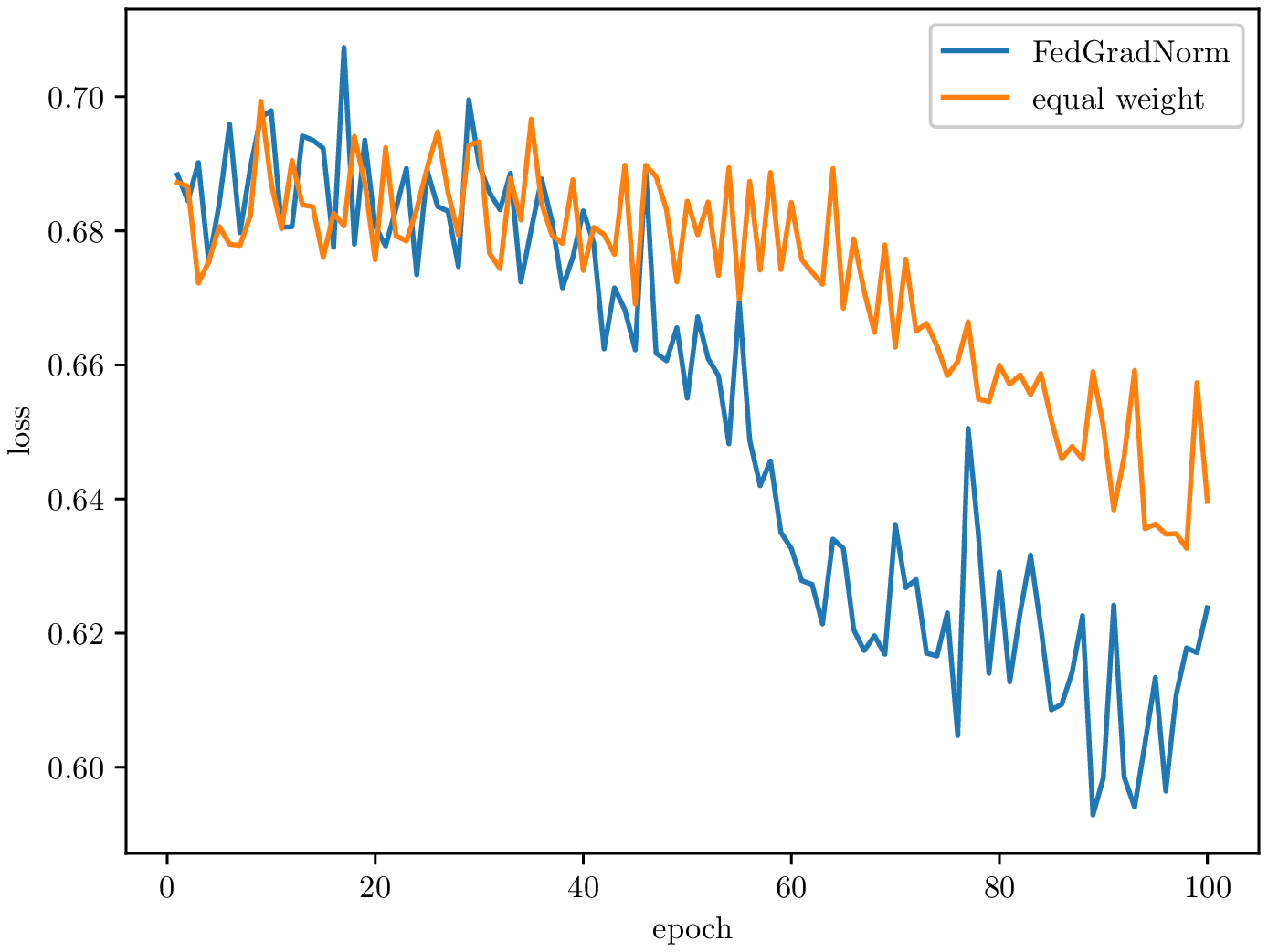}}\\ \vspace{-0.35cm}
 	\subfigure[]{%
 	\includegraphics[width=0.49\linewidth]{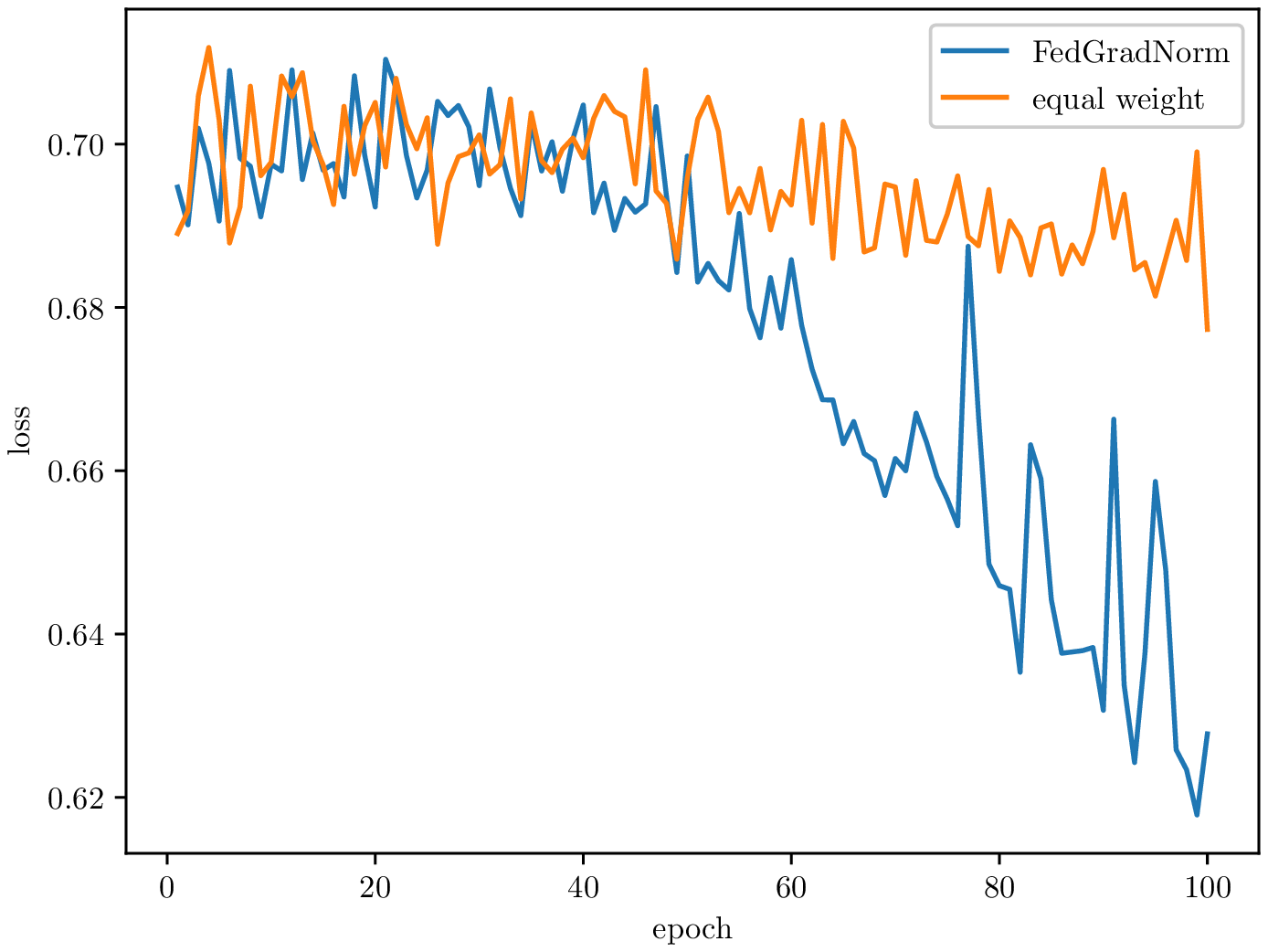}}
 	\subfigure[]{%
 	\includegraphics[width=0.49\linewidth]{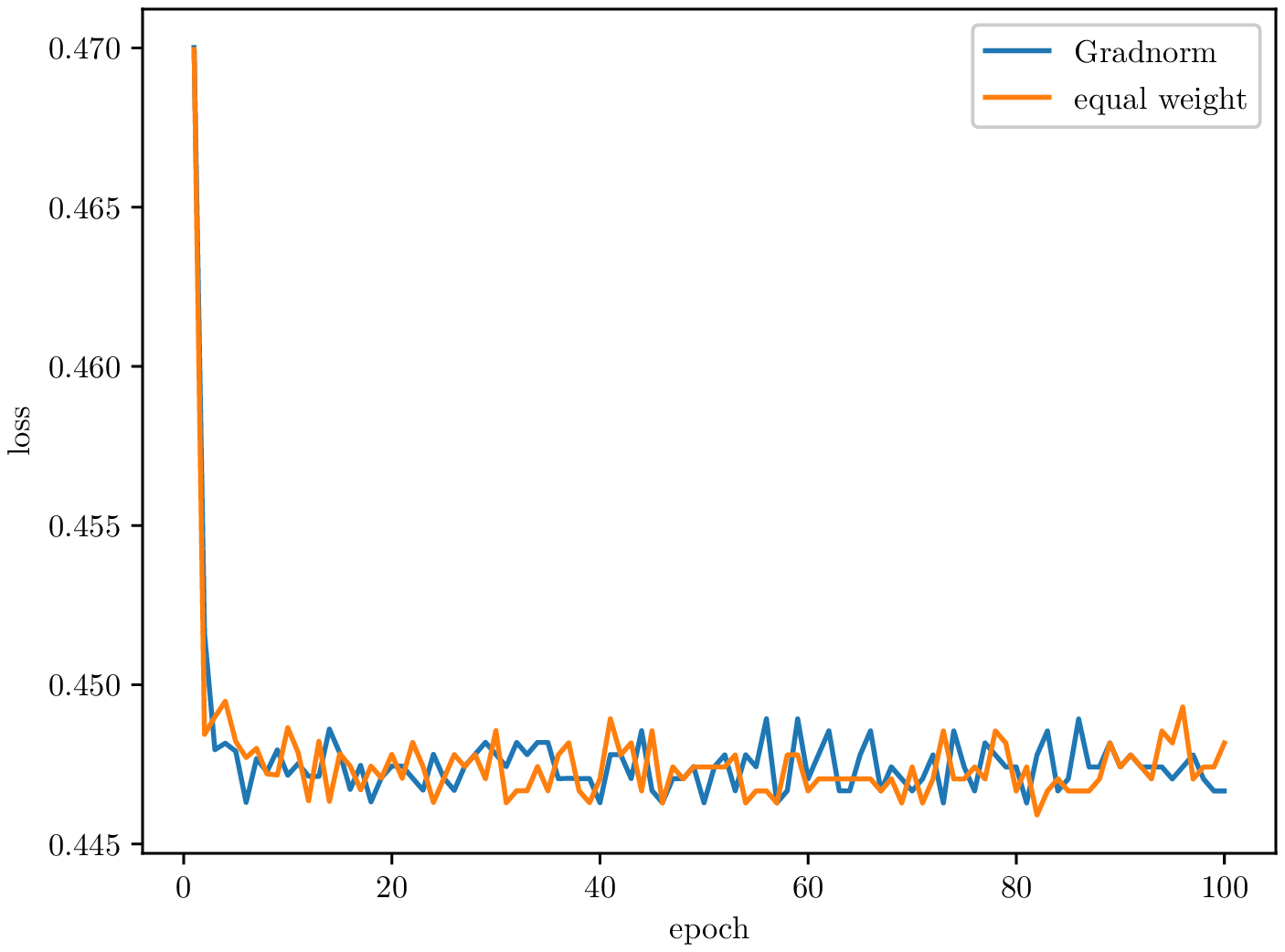}}\\ \vspace{-0.35cm}
 	\subfigure[]{%
 	\includegraphics[width=0.49\linewidth]{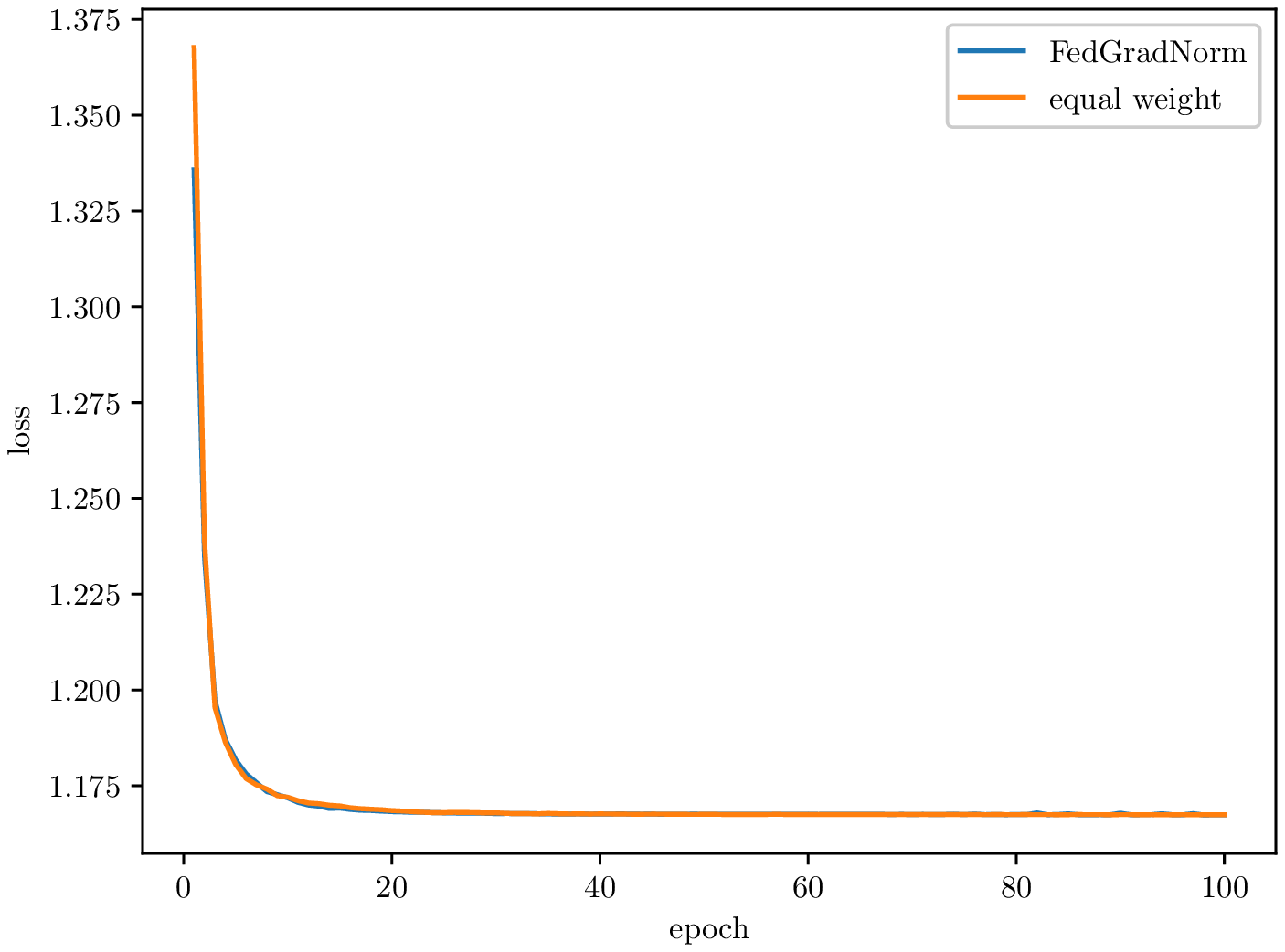}}
 	\subfigure[]{%
 	\includegraphics[width=0.49\linewidth]{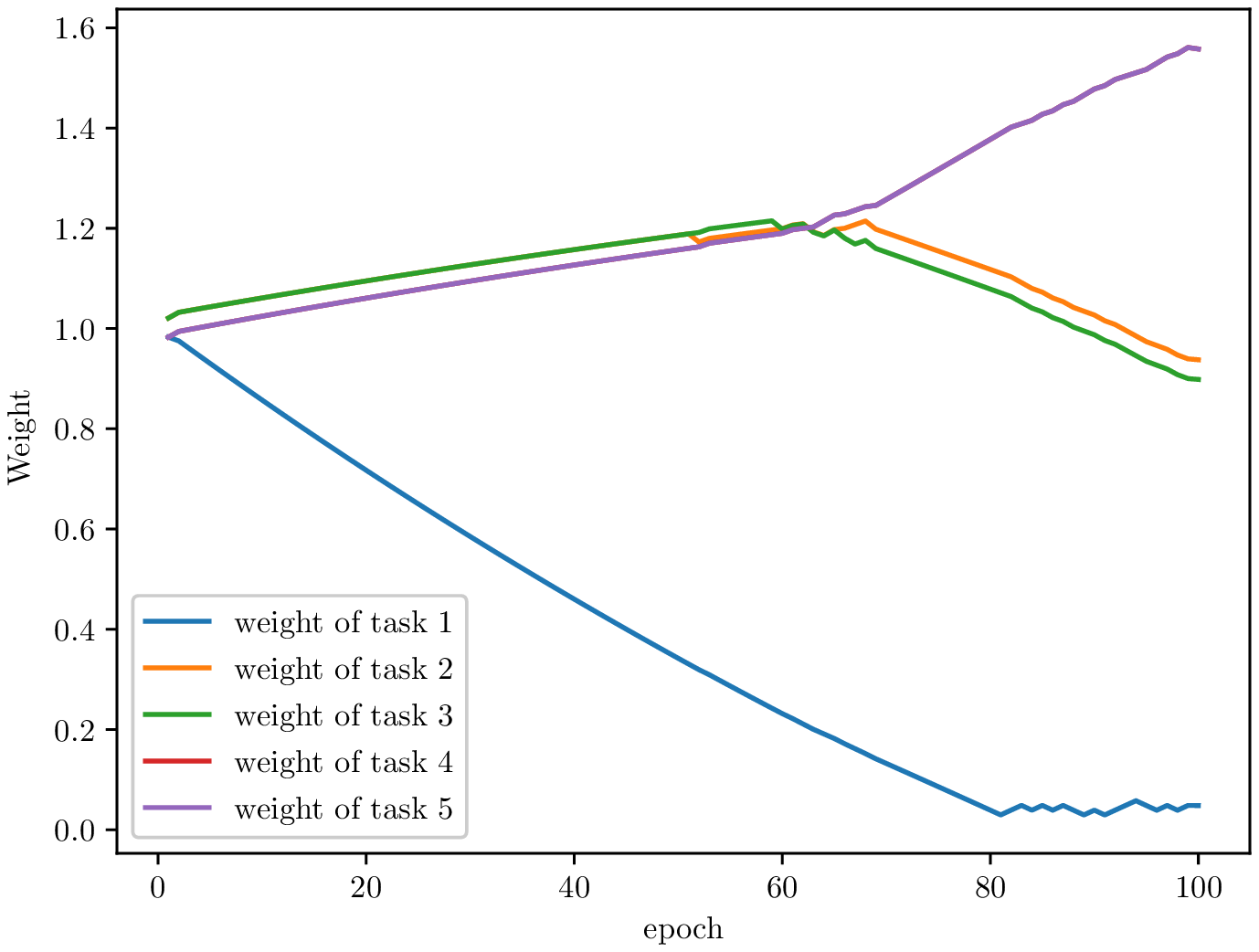}}
 	\end{center}
 	\vspace{-0.2cm}
 	\caption{Comparison of task losses in \emph{FedGradNorm} and \emph{FedRep}; balanced data allocation among tasks (a) task 1 (face landmark), (b) task 2 (gender), (c) task 3 (smile), (d) task 4 (glasses), (e) task 5 (pose), (f) task weights.}
 	\label{loss_gradnorm}
 	\vspace{-0.3cm}
\end{figure}

We first start with the MTFL dataset. Since the first task (determining the face landmarks) is a regression task, it has higher gradient than the other tasks which are all classification tasks. As shown in Fig.~\ref{loss_gradnorm}, \emph{FedGradNorm}, which is a dynamic-weighting method will gradually decrease the weight of the first task so that the other tasks can optimize their corresponding losses. Starting from epoch 70, when task 2 and 3 finally can decrease their loss with a higher rate, their corresponding task weights decrease to improve the two remaining tasks. Without using the dynamic-weighting method, tasks 2 and 3 could not be improved since the task 1 would mask the remaining tasks' gradient updates. Task 4, detecting glasses on human faces, reaches the minimum very fast at the first epoch since it is an easy task compared to the others. Therefore, its performance does not improve much, as shown in Fig.~\ref{loss_gradnorm}. Although the performance of tasks 1 and 5 are also quite the same in the long-run, \emph{FedGradNorm} helps to learn a bit faster at the early stages. For Fig.~\ref{loss_gradnorm}, the data allocation is balanced.

Next, we investigate the case where the data allocation is imbalanced, namely, some clients have a smaller portion of the dataset. In the following simulation, task 2 and task 4  have access to 500 data points while other tasks have 3000 data points to use in the training procedure. As shown in Table~\ref{tab:loss_gradnorm_imbalance}, \emph{FedGradNorm} again has a better performance compared to the equal-weighting case in \emph{FedRep}.

\begin{table}[H]
    \begin{center}
    \begin{tabular}{|c|c|c|c|c|c|}
        \hline
        Tasks & face landmark & gender & smile & glass & pose \\
        \hline
        \emph{FedRep} loss & 33.28 & 0.66 & 0.60 & 0.44 & 1.1\\
        \hline
        \emph{FedGradNorm} loss & 33.25 & \textbf{0.56} & \textbf{0.57} & 0.43 & 1.1\\
        \hline
    \end{tabular}
    \end{center}
    \caption{Comparison of task losses after 100 epochs in \emph{FedGradNorm} and \emph{FedRep}; imbalanced data allocation among tasks.}
    \label{tab:loss_gradnorm_imbalance}
    \vspace*{-0.3cm}
\end{table}

\begin{figure}[]
 	\begin{center}
 	\subfigure[]{%
 	\includegraphics[width=0.49\linewidth]{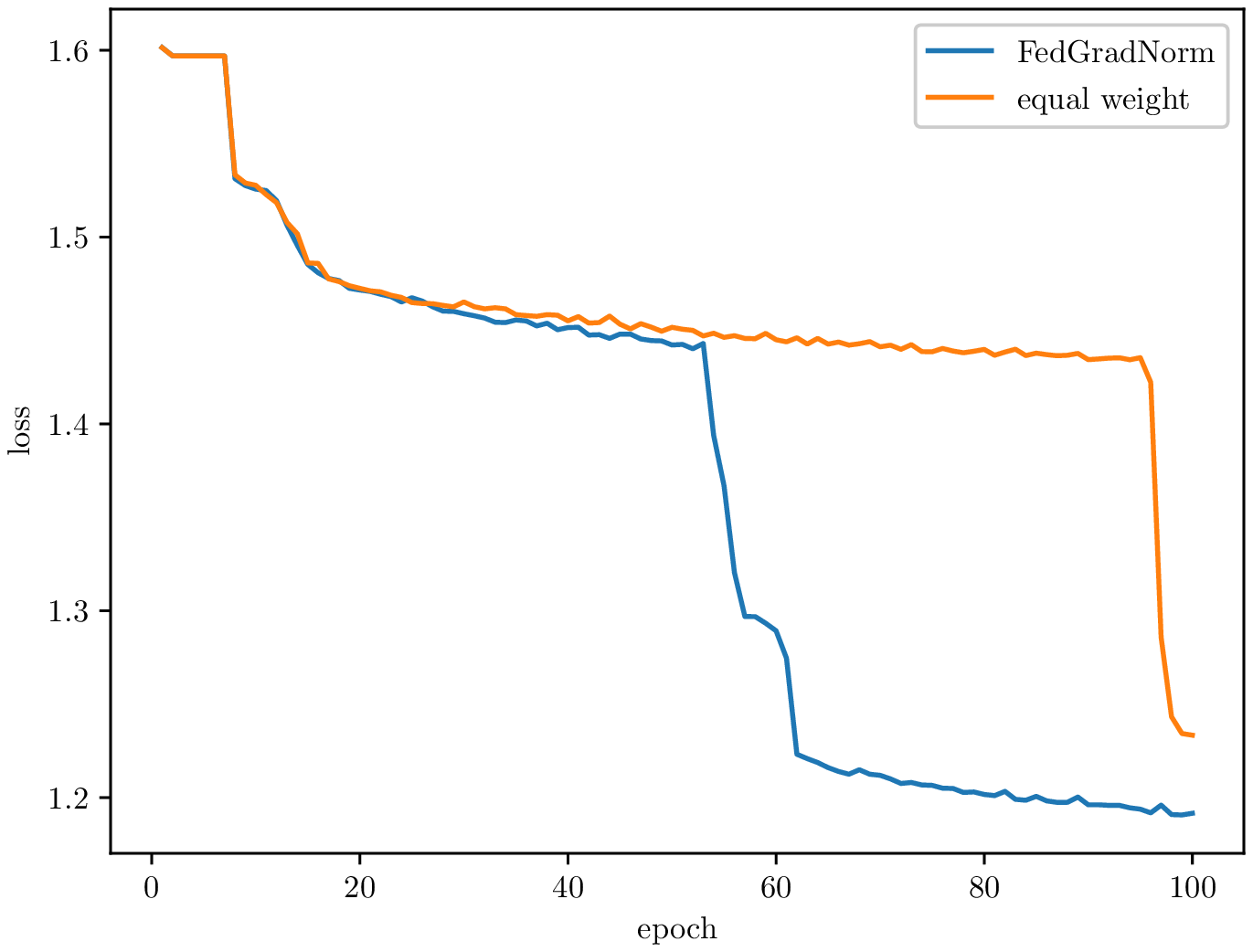}}
 	\subfigure[]{%
 	\includegraphics[width=0.49\linewidth]{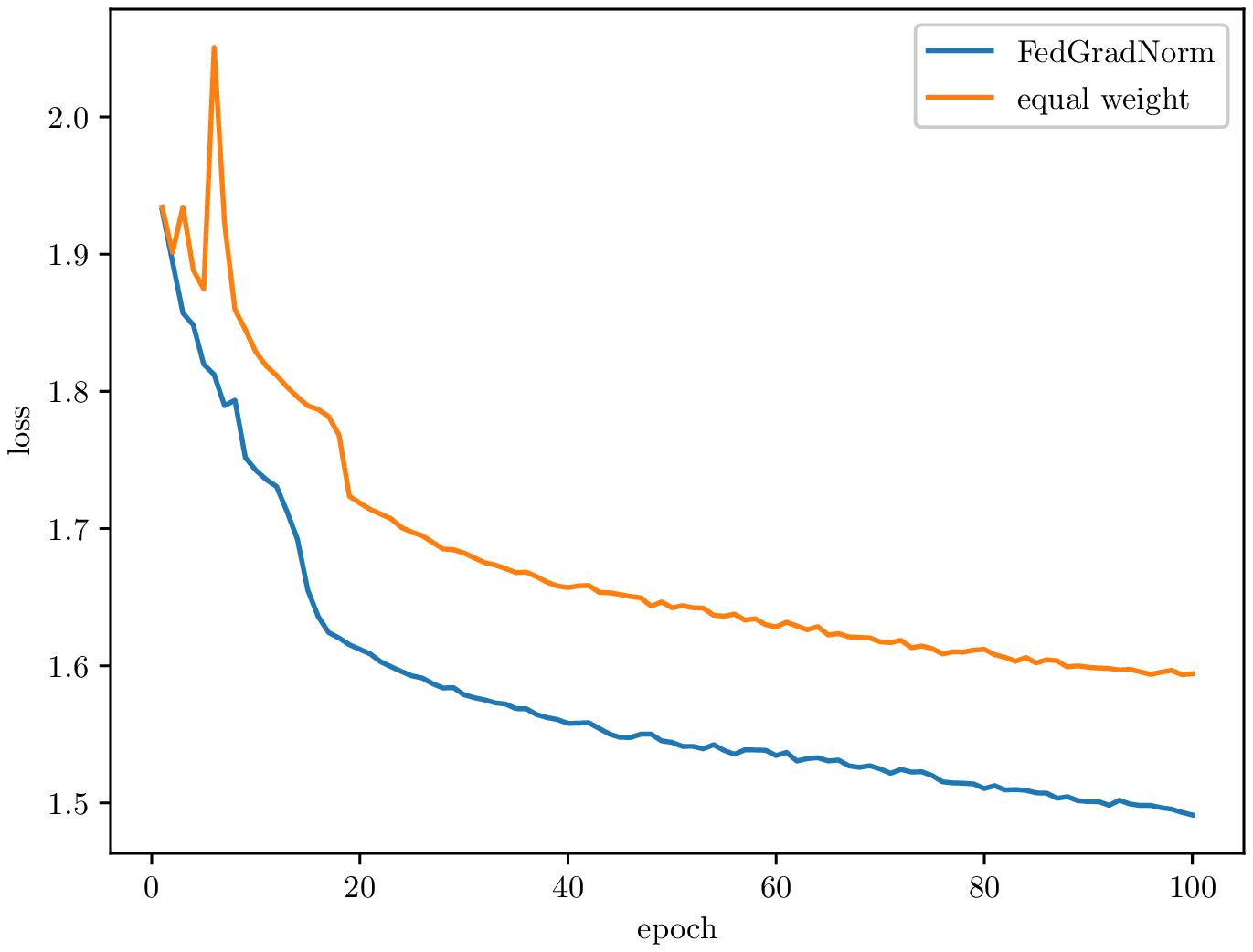}}\\ \vspace{-0.3cm}
 	\subfigure[]{%
 	\includegraphics[width=0.49\linewidth]{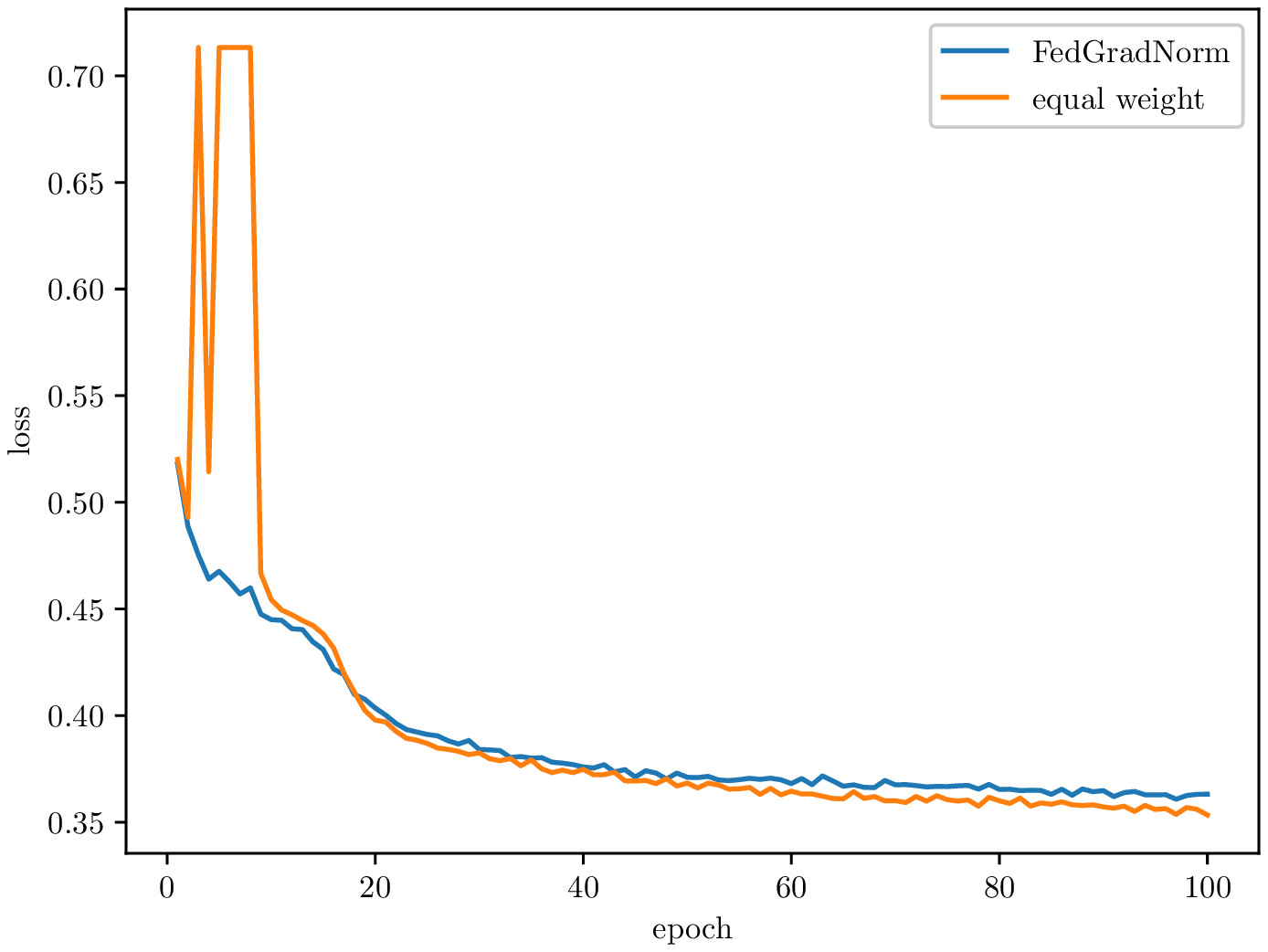}}
 	\subfigure[]{%
 	\includegraphics[width=0.49\linewidth]{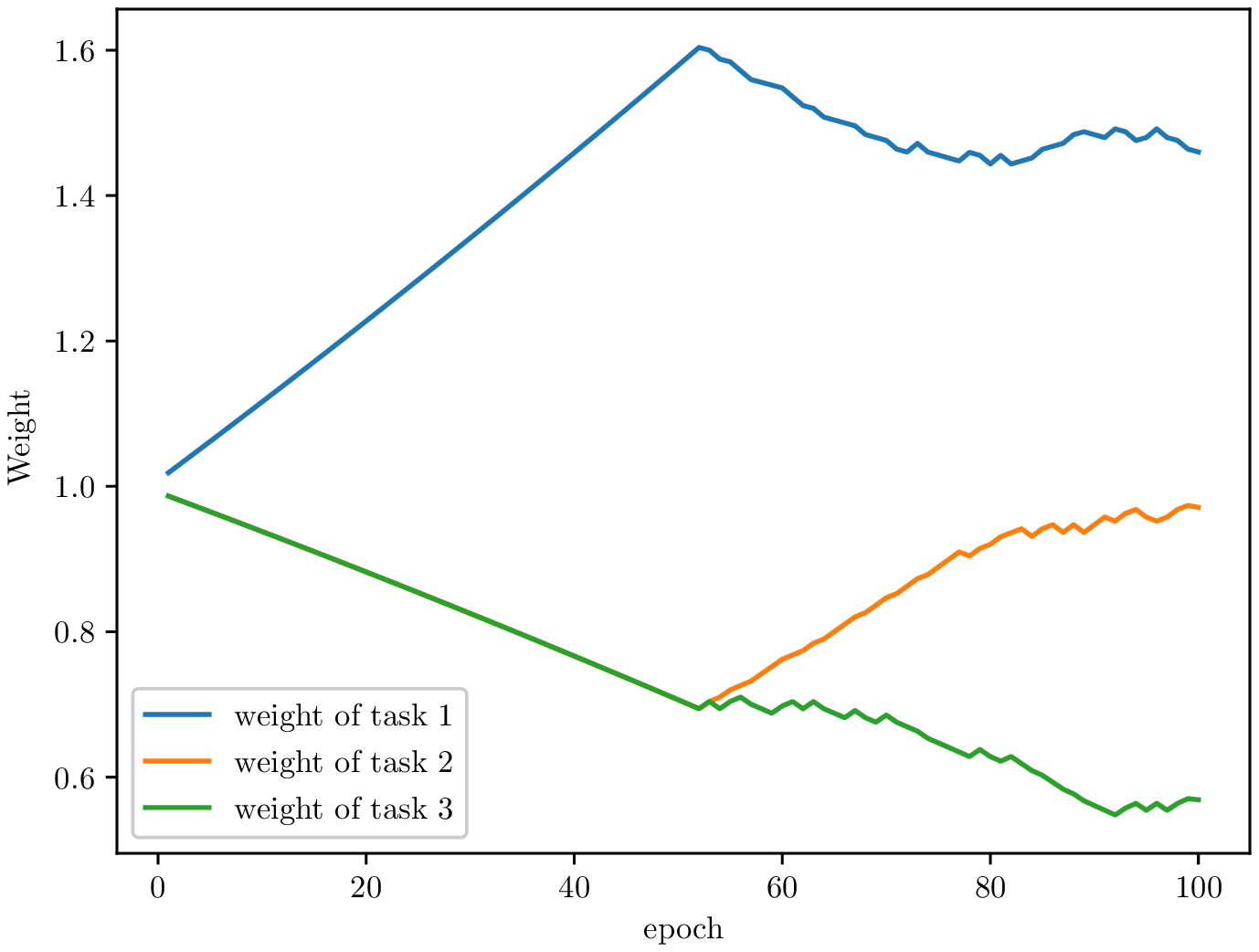}}
 	\end{center}
 	\caption{Comparison between task accuracy achieved via \emph{FedGradNorm} and \emph{FedRep} in RadComDynamic dataset (a) task 1 (modulation classification), (b) task 2 (signal classification), (c) task 3 (anomaly behavior), (d) task weights.}
 	\label{loss_Fedgradnorm_wireless}
 	\vspace{-0.3cm}
\end{figure}

Next, we consider the RadComDynamic dataset by using Network 2. As shown in Fig.~\ref{loss_Fedgradnorm_wireless}, the result again indicates the superiority of \emph{FedGradNorm} compared to \emph{FedRep} on modulation detection and signal detection tasks. Based on the loss value, task 1 (modulation detection task) and task 2 (signal detection task) are harder and slower to learn than the anomaly detection task. As the result shows, using the dynamic-weighting based \emph{FedGradNorm} can ensure that the signal and modulation detection classes, which are slower in training, have the same opportunity as the other task to improve their performance. Also, since the loss of task 2 and task 3 decreases with the same and higher slope at the beginning stage than the task 1, the dynamic-weighting method increases the corresponding task weight for the task 1 to push the task to be trained faster. In epoch 55, when the loss of task 1 decreases significantly, the weight of task 1 decreases to let other tasks to be trained more strongly. 

\bibliographystyle{unsrt}
\bibliography{reference}
\end{document}